\documentclass{article}
\usepackage{ijcai24}

\usepackage{times}
\usepackage{soul}
\usepackage{url}
\usepackage[hidelinks]{hyperref}
\usepackage[utf8]{inputenc}
\usepackage[small]{caption}
\usepackage{graphicx}
\usepackage{amsmath}
\usepackage{amsthm}
\usepackage{booktabs}
\usepackage{algorithm}
\usepackage{algorithmic}
\usepackage[switch]{lineno}

\usepackage{xcolor}         
\usepackage{colortbl}
\usepackage{multirow}
\usepackage{amssymb}
\usepackage{makecell}
\usepackage{tikz}
\usepackage{caption}
\usepackage{newfloat}
\usepackage{listings}
\usepackage{comment}

\title{Masked Generative Extractor for Synergistic Representation and \\3D Generation of Point Clouds}

\author{
Hongliang Zeng
\and
Ping Zhang\thanks{Corresponding author}
\and
Fang Li\and
Jiahua Wang\and
Tingyu Ye\And
Pengteng Guo
\affiliations
South China University of Technology, Guangzhou, China
\emails
pzhang@scut.edu.cn
}

\begin{document}

\maketitle

\begin{abstract}
	Representation and generative learning, as reconstruction-based methods, have demonstrated their potential for mutual reinforcement across various domains. In the field of point cloud processing, although existing studies have adopted training strategies from generative models to enhance representational capabilities, these methods are limited by their inability to genuinely generate 3D shapes. To explore the benefits of deeply integrating 3D representation learning and generative learning, we propose an innovative framework called \textit{Point-MGE}. Specifically, this framework first utilizes a vector quantized variational autoencoder to reconstruct a neural field representation of 3D shapes, thereby learning discrete semantic features of point patches. Subsequently, we design a sliding masking ratios to smooth the transition from representation learning to generative learning.  Moreover, our method demonstrates strong generalization capability in learning high-capacity models, achieving new state-of-the-art performance across multiple downstream tasks. In shape classification, Point-MGE achieved an accuracy of 94.2\% (+1.0\%) on the ModelNet40 dataset and 92.9\% (+5.5\%) on the ScanObjectNN dataset. Experimental results also confirmed that Point-MGE can generate high-quality 3D shapes in both unconditional and conditional settings.
\end{abstract}

\section{Introduction}
Recent studies have confirmed that masked modeling-based representation learning frameworks~\cite{Yu_2022_CVPR,pang2022masked,zhang2022point} can extract high-quality 3D representations. At the same time, generative modelss~\cite{zhang20223dilg,Mittal_2022_CVPR,zhang20233dshape2vecset} have achieved remarkable success in generating realistic 3D shapes. Nevertheless, few studies have explored the deep integration of these two models to unlock their combined potential in 3D shape processing. In natural language processing (NLP)~\cite{touvron2023llama} and 2D image domains~\cite{Li_2023_CVPR}, we have witnessed the versatility of pre-trained models: not only can they be used for diverse generative tasks such as question answering (QA), code generation, and image generation, but they can also serve as feature extractors for various downstream tasks through fine-tuning. These achievements prompt us to consider: can the benefits of this model integration be similarly applied to the representation of point cloud data and 3D shape generation? To this end, we conducted an in-depth analysis of the characteristics of both models and found that masked reconstruction methods offer a potential pathway to achieve this integration, though several challenges remain.

Firstly, generative learning is easily trained on discrete distributions, whereas existing point cloud representation learning~\cite{pang2022masked,zhang2022point} tends to operate within continuous feature spaces. In prior work, Point-Bert~\cite{Yu_2022_CVPR} learns a tokenizer by reconstructing the input point cloud and optimizing the chamfer distance (CD) loss. However, using point clouds directly as the reconstruction target introduces a sampling bias. This is because 3D shapes are inherently continuous, but point clouds sampled from these shapes can only approximate them in a discrete form. This sampling bias can lead to non-zero loss differences between different point cloud samples of the same object, thereby affecting the precise understanding of shape features. Moreover, methods that rely solely on point cloud reconstruction are limited in capturing high-level representations of 3D shapes. For instance, they cannot generate richer shape descriptions such as neural radiance fields (NeRFs) or triangular meshes, which restricts the model's potential in generative tasks. To address this, we shift the reconstruction target to 3D shapes represented by NeRFs and utilize a vector quantized variational autoencoder (VQVAE)~\cite{van2017neural} to extract discrete semantic features from point clouds. The Point-MGE framework leverages these high-level semantic discrete features not only as input but also as reconstruction targets, thereby learning the probability distribution of masked tokens. During the vector quantization process, different point cloud samples from the same local region are consistently mapped to the same discrete tokens, ensuring consistency and robustness in the representations.

Secondly, the inherent unordered nature of point cloud poses a challenge for Transformer-based generative models, as these models require serialized inputs. To address this issue, we partition the complete point cloud into multiple patches, each defined by a fixed number of points and a central point. To serialize these patches, we adopt the Morton sorting algorithm used in PointGPT~\cite{NEURIPS2023_5ed5c3c8}, which can sort point cloud patches while preserving geometric relationships. In generation task, since the central point of point cloud patch is unknown, we introduce relative position encoding in the generator of Point-MGE as a replacement for traditional absolute position encoding, and use the central points as the reconstruction target. 

Finally, generative training tends to adopt higher masking ratios, or even full masking, whereas representation learning typically uses relatively low masking ratios. In 2D image processing, MAGE~\cite{Li_2023_CVPR} achieves variable masking ratios by randomly sampling them from a Gaussian distribution, utilizing these variable ratios to perform the task. However, due to the irregularity of point cloud data and the lack of support from large-scale datasets, such drastic fluctuations in masking ratios may adversely affect model convergence. To address this issue, we have carefully designed a sliding masking ratio mechanism to smoothly adjust the masking ratio during training. Specifically, the masking ratio gradually increases as training iterations progress, facilitating a smooth transition from representation learning to generative training. We also conducted a comparative analysis of the effects of different sliding curves. Point-MGE can scale to high-capacity model training and has demonstrated outstanding performance across multiple benchmark datasets. It achieved an accuracy of 94.2\% on the ModelNet40 dataset~\cite{Wu_2015_CVPR} for shape classification and 92.9\% on the ScanObjectNN dataset~\cite{Uy_2019_ICCV}. Additionally, in the part segmentation task on the ShapeNetPart dataset~\cite{yi2016scalable}, our method attained an average class Intersection over Union (IoU) of 85.0\%. Furthermore, we also showcased the capability of the pre-trained model to generate realistic 3D shapes in both unconditional and conditional settings.

Our main contributions can be summarized as follows: (I) We propose Point-MGE, a unified framework for point cloud data representation learning and generative training that combines a two-stage training method with a sliding masking ratio design. (II) Point-MGE overcomes the variance issue in point cloud sampling and scales to high-capacity model training, setting a new SOTA benchmark for downstream tasks. (III) Point-MGE is capable of generating high-quality 3D shapes in both unconditional and conditional settings. 

\section{Related Works}
\paragraph{Representation Learning for Point Clouds.}
In recent years, self-supervised frameworks for point cloud representation learning have made significant progress, forming two main model paradigms. Firstly, methods based on contrastive learning~\cite{xie2020pointcontrast,Zhang_2021_ICCV,Afham_2022_CVPR} have played an important role in point cloud self-supervised learning. These methods can explore the intrinsic properties of point clouds, but they often rely on cross-modal information or teacher models, which undoubtedly add extra burden to the training process. Secondly, inspired by natural language processing~\cite{devlin2018bert,brown2020language} and 2D computer vision~\cite{he2022masked,Wei_2022_CVPR}, a more closely related area to our work is another category of methods based on masked point cloud modeling~\cite{Yu_2022_CVPR,pang2022masked,zhang2022point,NEURIPS2023_5ed5c3c8}. Point-M2AE~\cite{zhang2022point} proposed an innovative hierarchical transformer structure, extending Point-MAE~\cite{pang2022masked} to capture multi-scale features of point clouds. On the other hand, PointGPT~\cite{NEURIPS2023_5ed5c3c8} attempts to solve the problem of global shape leakage through autoregressive generation, but it does not have the ability to generate high-quality 3D models from scratch. Additionally, methods based on masked region reconstruction are susceptible to point cloud sampling variance, which may introduce additional understanding and bias issues. To overcome these challenges, our method adopts a strategy of masked modeling in the high-level semantic space, effectively addressing the aforementioned issues.

\paragraph{Generative Models for 3D Shapes.}
In the field of 3D generation, generative adversarial networks (GANs)~\cite{achlioptas2018learning,Chen_2019_CVPR,Ibing_2021_CVPR,zheng2022sdf} constitute a major generative paradigm, despite their limitations in training stability and mode collapse issues. Meanwhile, some research has turned to alternative approaches such as normalizing flows (NFs)~\cite{dinh2014nice,rezende2015variational,yang2019pointflow}, energy-based models (EBMs)~\cite{xie2016theory,xie2020generative,Xie_2021_CVPR}, and autoregressive models (ARs)~\cite{Mittal_2022_CVPR,cheng2022autoregressive,Yan_2022_CVPR,zhang20223dilg}. Recently, inspired by the significant progress of diffusion models (DMs)~\cite{dhariwal2021diffusion} in image synthesis, some studies have begun to explore the application of diffusion models to 3D shape generation~\cite{zhou20213d,vahdat2022lion,hui2022neural,Shue_2023_CVPR,Chou_2023_ICCV,zhang20233dshape2vecset} . Although these models can generate realistic 3D models, they often lack the ability to extract high-quality semantic representations from 3D shapes, limiting their direct application in point cloud representation learning tasks. To address this challenge, we propose an innovative approach aimed at generating diverse samples while extracting high-quality semantic embeddings from 3D shapes to promote data generalization and enhance the performance of downstream tasks. 

\section{Method}
\begin{figure*}[!t]
	\centering
	\includegraphics[width=0.96\linewidth,keepaspectratio]{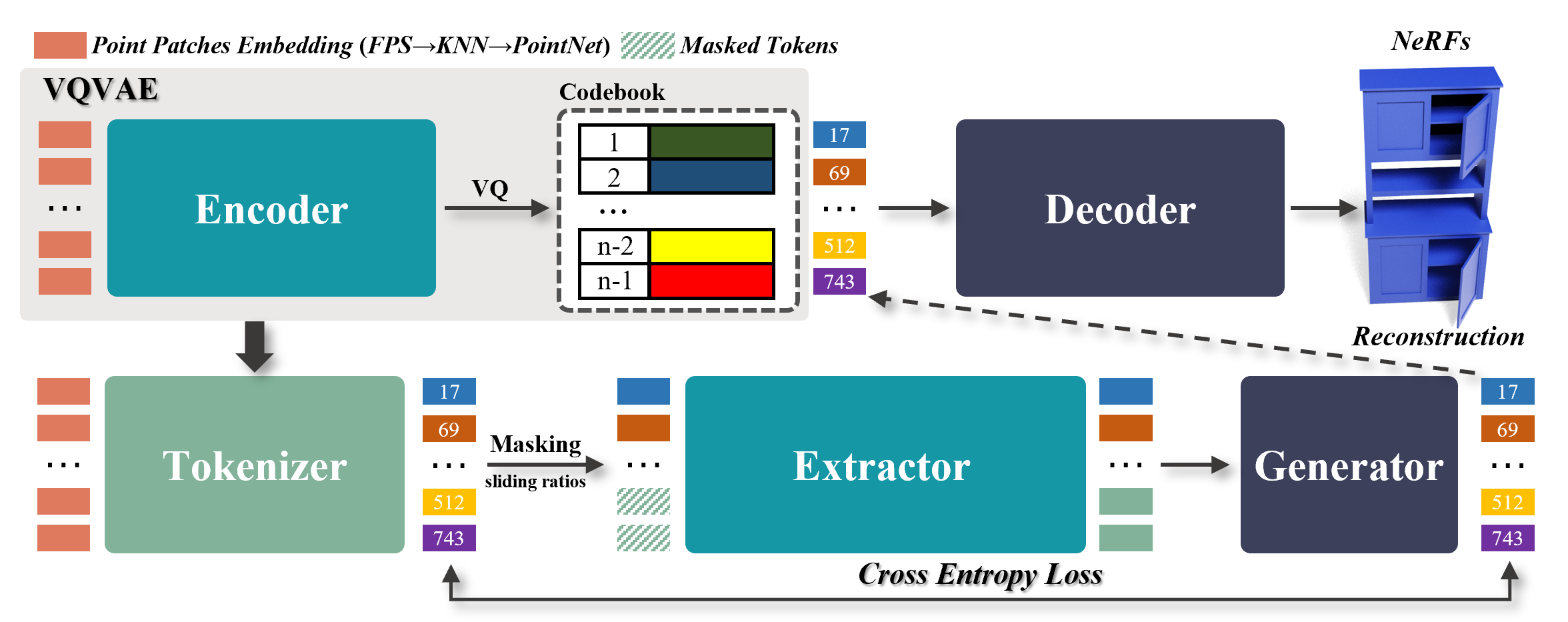}
	\caption{\textbf{Overall pipeline of Point-MGE.} First, a VQVAE is used to reconstruct 3D shapes represented by NeRFs, converting the input point cloud into a series of discrete semantic tokens. Then, where a ViT-based extractor-generator architecture is employed to extract high-quality feature representations from the unmasked tokens and to reconstruct the masked tokens.}
	\label{pipeline}
\end{figure*}

Point-MGE is designed as a unified framework for both 3D generation tasks and representation learning. As shown in Figure~\ref{pipeline}, the framework first divides the input point cloud into multiple local point patches and serializes them based on their geometric positional relationships. Then, it uses a mini-PointNet~\cite{Qi_2017_CVPR} model for initial feature extraction of each point patch. Subsequently, by training a VQVAE model, these features are quantized into discrete tokens. Finally, the framework combines a variable masking ratio with a vision transformer (ViT)-based extractor-generator architecture to predict masked tokens.

\subsection{Point Patch Embedding}
\paragraph{Point Cloud Patching.}
We determine the number of patches and the number of points per patch in the point cloud sampling process through two key hyperparameters $G$ and $K$. Specifically, given the input point cloud $X \in \mathbb{R} ^ {n \times 3}$, we use the farthest point sampling (FPS) strategy to select $G$ central points,
\begin{equation}
	C = \{x_i\}_{i \in \mathcal{N}} = \text{FPS}\left(X\right),~C \in \mathbb{R} ^ {G \times 3},
\end{equation}
where $|\mathcal{N}| = n$. Then, using the K-nearest neighbors (KNN) algorithm, we identify $|\mathcal{N}_i| = K$ neighbors of the central point $\{x_i\}_{i \in \mathcal{N}}$, resulting in $G$ point cloud patches,
\begin{equation}
	P = \{x_j\}_{j \in \mathcal{N}_i} = \text{KNN}\left(C, X\right),~P \in \mathbb{R} ^ {G \times K \times 3}.
\end{equation}

\paragraph{Embedding.}
To provide serialized input, we follow~\cite{NEURIPS2023_5ed5c3c8} and encode the coordinates of the central points $x_i$ into one-dimensional space using morton ordering~\cite{morton1966computer}, and then sort them to obtain serialized central points $C^s \in \mathbb{R} ^ {G \times 3} $ and point cloud patches $P^s \in \mathbb{R} ^ {G \times 3}$. We project each patch into an embedding vector using the mini-PointNet\cite{Qi_2017_CVPR} module,
\begin{equation}
	P^s_E = \{e_i\}_{i \in \mathcal{G}} = \text{PointNet}\left(P^s\right),~P^s_E \in \mathbb{R} ^ {G \times D},
\end{equation}
where $|\mathcal{G}| = G$, D is the embedding dimension of the point patch.

\subsection{Tokenization}
We train a VQVAE~\cite{van2017neural} to obtain a quantizer that embeds input point patches into a series of semantic tokens. This tokenization step enables our model to focus on high-level semantics, which helps to overcome the variance issue in point cloud sampling and allows for a more comprehensive understanding of 3D shapes.

\paragraph{Vector Quantization.}
We use a ViT-based encoder to learn the local latents $T$. The input to the transformer includes the sorted point block embeddings $P^s_E$ and the sinusoidal positional encodings~\cite{vaswani2017attention} $C^s_E$ of the center point,
\begin{equation}
	T = \{t_i\}_{i \in \mathcal{G}} = \text{Enc}\left(P^s_E, C^s_E\right),~T \in \mathbb{R} ^ {G \times D}.
\end{equation}
The output of the encoder is then replaced with the closest vector in the codebook $\mathcal{Q}$ through vector quantization,
\begin{equation}
	T^Q = \{t^q_i\}_{i \in \mathcal{G}} = \mathop{\arg\min}\limits_{t^q_j \in \mathcal{Q}} \Vert t^q_j, t_i\Vert,~T^Q \in \mathbb{R} ^ {G \times D}.
\end{equation}

\paragraph{Reconstruction.}
We utilize a ViT-based decoder to reconstruct the 3D shape's NeRFs $\hat{\mathcal{O}}(x) \in \{0,1\}$, where $\hat{\mathcal{O}}(x) = 1$ indicates that the point $x$ in space belongs to the object. The decoder takes as input the quantized tokens and the position encoding of the center points,
\begin{equation}
	T^D = \{t^d_i\}_{i \in \mathcal{G}} = \text{Dec}\left(T^Q, C^s_E\right),~T^D \in \mathbb{R} ^ {G \times D}.
\end{equation}
For any $x \in \mathbb{R}^3 $, we use the nadaraya-watson estimator for interpolation to obtain its latent $t^d_x$. Then, we estimate $\mathcal{O}(x)$ through a multi-layer perceptron and a sigmoid activation function,
\begin{equation}
	\mathcal{O}(x) = \text{Sigmoid}\left(\text{MLP}\left(x, t^d_x\right)\right).
\end{equation}
The optimization objective consists of binary cross-entropy (BCE) loss and an additional commitment loss,
\begin{equation}
	\footnotesize
	\begin{split}
		\mathcal{L}_{n} = \mathbb{E}_{x \in \mathbb{R}^3}[\text{BCE}(\mathcal{O}(x), \hat{\mathcal{O}}(x))] +
		~~~~~~~~~~~~\\ \lambda\mathbb{E}_{x \in \mathbb{R}^3}[\mathbb{E}_{i \in \mathcal{M}}\Vert \epsilon(t^q_i) - t_i \Vert^2],
	\end{split}
\end{equation}
where $\epsilon(\cdot)$ represents the stop-gradient operation. In this paper, \(\lambda = 0.25\) is a weighting coefficient.

\subsection{Pre-training}
After obtaining the tokenizer, we simultaneously train an extractor and a generator model through masking modeling. Below, we introduce the masking strategy, model design, and optimization objectives of Point-MGE.

\begin{table*}[!t]
	\centering
	\footnotesize
	\setlength{\tabcolsep}{6pt}
	\begin{tabular}{lccccccc}
		\toprule
		\multirow{2}{*}{Methods} &\multirow{2}{*}{Reference} &\multirow{2}{*}{Params(M)} &\multicolumn{2}{c}{ModelNet40} &\multicolumn{3}{c}{ScanObjectNN}\\
		\cmidrule(r){4-5} \cmidrule(r){6-8}
		& & &w/o Vote &w/ Vote &OBJ\_BG &OBJ\_ONLY &PB\_T50\_RS\\
		\midrule
		\multicolumn{8}{c}{\textit{Supervised Learning Only}}\\
		\midrule
		PointNet~\cite{Qi_2017_CVPR} &CVPR 2017 &3.5 &- &89.2  &73.3 &79.2 &68.0\\
		PointNet++~\cite{qi2017pointnet++} &Neurips 2017 &1.5 &- &90.7 &82.3 &84.3 &77.9\\
		DGCNN~\cite{wang2019dynamic} &TOG 2019 &1.8 &- &92.9 &82.8 &86.2 &78.1\\
		PointMLP~\cite{ma2022rethinking} &ICLR2022 &12.6 &94.1 &94.5 &- &- &85.2\\ 
		P2P-HorNet~\cite{wang2022p2p} &NeurIPS 2022 &195.8 &94.0 &- &- &- &89.3\\
		\midrule
		\multicolumn{8}{c}{\textit{with Single-Modal Pre-Training}}\\
		\midrule
		Point-BERT~\cite{Yu_2022_CVPR} &CVPR 2022 &22.1 &92.7 &93.2 &87.4 &88.1 &83.1\\
		Mask-Point~\cite{liu2022masked} &ECCV 2022 &22.1 &- &93.8 &89.3 &88.1 &84.3 \\
		Point-MAE~\cite{pang2022masked}  &ECCV 2022 &22.1 &93.2 &93.8 &90.0 &88.2 &85.2 \\
		Point-M2AE~\cite{zhang2022point} &Neurips 2022 &15.3 &93.4 &94.0 &91.2 &88.8 &86.4 \\
		PointGPT~\cite{NEURIPS2023_5ed5c3c8} &Neurips 2023 &19.5 &- &94.0 &91.6 &90.0 &86.9 \\
		GPM~\cite{li2024general} &CVPR2024 &41.5 &\textbf{94.0} &\textbf{94.3} &90.2 &90.0 &84.8 \\
		\textbf{Point-MGE} &- &15.3 &93.8 &94.2 &\textbf{92.9} &\textbf{91.0} &\textbf{87.1} \\
		\textit{Improvement (baseline: Point-BERT)} &-&-&\textit{+1.1} &\textit{+1.0} &\textit{+5.5} &\textit{+2.9} &\textit{+4.0}\\
		\midrule
		Point-MAE-B~\cite{pang2022masked}  &ECCV 2022 &120.1 &- &94.2 &94.2 &93.9 &90.2 \\
		Point-M2AE-B~\cite{zhang2022point} &Neurips 2022 &77.5 &- &94.3 &95.2 &94.3 &91.2 \\
		PointGPT-B~\cite{NEURIPS2023_5ed5c3c8} &Neurips 2023 &82.1 &- &94.4 &95.8 &95.2 &91.9 \\
		\textbf{Point-MGE-B} &- &77.5 &- &\textbf{94.5} &\textbf{96.1} &\textbf{95.7} &\textbf{92.3} \\
		\midrule
		\multicolumn{8}{c}{\textit{with Cross-Modal Pre-Training}}\\
		\midrule
		ACT~\cite{dong2023autoencoders} &ICLR 2023 &22.1 &93.2 &93.7 &93.3 &91.9 &88.2 \\
		Joint-MAE~\cite{guo2023joint} &IJCAI 2023 &- &- &94.0 &90.9 &88.9 &86.1 \\
		I2P-MAE~\cite{zhang2023learning} &CVPR 2023 &15.3 &93.7 &94.1 &94.2 &91.6 &90.1 \\
		Recon~\cite{qi2023contrast} &ICML 2023 &44.3 &94.1 &94.5 &95.2 &93.3 &90.6\\
		\bottomrule
	\end{tabular}
	\caption{\textbf{Shape classification results on ModelNet40~\protect\cite{Wu_2015_CVPR} and ScanObjectNN~\protect\cite{Uy_2019_ICCV}.} We report the accuracy (\%) of both training from scratch and fine-tuning models on ModelNet40 and ScanObjectNN datasets.}
	\label{Shape Classification}
\end{table*}

\paragraph{Masking Strategy.}
For a given point cloud $X \in \mathbb{R}^{n \times 3}$, we first use the trained tokenizer to obtain the corresponding discrete tokens $T^Q$. Subsequently, we introduce a sliding masking ratio to apply masking, which can be calculated as:
\begin{equation}
	m_r = \beta + (1 - \beta)(\frac{\gamma}{\Gamma})^{u},~m_r \in \left[\beta,1\right],
\end{equation}
where, $\gamma$ and $\Gamma$ denote the current training iteration and the total number, respectively. The parameter $u > 0$ is a hyperparameter used to control the trend of the sliding function. After masking, to reduce training time and memory consumption, we randomly drop a portion of masked tokens, ensuring that only $V = (1-\beta)\times G$ tokens are input to the extractor.

\paragraph{Extractor and Generator.}
We feed the remaining tokens into the ViT-based extractor-generator structure. Then, we concatenate a learnable mask token to the input sequence. The extractor processes the input tokens $T^Q_m$ and encodes them into the latent feature space,
\begin{equation}
	\footnotesize
	T^E = \{t^e_i\}_{i \in \mathcal{G}, m_i=0} = \text{Ext}(T^Q_m, C^s_E),~T^E \in \mathbb{R}^{V \times D},
\end{equation}
where $M = [m_i]^G_{i=1}$ represents a binary mask indicating whether a token is masked or not. Then, the output of the extractor is padded to the complete length $G$ as the input $T^E_p$ of the generator using the learnable mask token,
\begin{equation}
	T^G = \{t^g_i\}_{i \in \mathcal{G}} = \text{Gen}(T^E_p, C^s_{RE}, \mathcal{C}),~T^G \in \mathbb{R}^{G \times D},
\end{equation}
where $C^s_{RE}$ is the relative position encoding of the sorted center points $C^s$ to adapt to the needs of the generation task, and $\mathcal{C}$ is optional conditional information (e.g., category, image, and text).

\paragraph{Reconstructive Training.}
Our training target is to reconstruct the masked tokens along with the center point coordinates $\hat{c}_i$. The reconstruction of center points serves the generation task because our VQVAE decoder requires absolute position encoding as input to reconstruct 3D shapes. Our loss function consists of two parts: the cross-entropy loss for token reconstruction and the mean square error loss for center point reconstruction, which can be expressed as:
\begin{equation}
	\footnotesize
	\mathcal{L}_{rt} = -\mathbb{E}_{t^q_i \in \mathcal{Q}}[\mathop{\sum}\limits_{\forall i ,m_i = 1} \log p(t^q_i \vert T^G_m)] + \frac{1}{\vert M\vert}\mathop{\sum}\limits_{\forall i ,m_i = 1} \Vert c_i, \hat{c}_i \Vert^2,
\end{equation}

where $T^G_m$ represents the masked portion of the tokens outputted by the generator. 

\subsection{Iterative Generation}
After pretraining, following~\cite{Li_2023_CVPR}, we employ an iterative generation strategy to generate diverse 3D shapes. The generation process starts with all tokens being considered as masked tokens. Using a cosine function, we determine the number of masked tokens to be replaced in each iteration. Specifically, in each iteration, based on the predicted probabilities from the generator, we sample and replace those tokens with higher probabilities.

\section{Experiments}
In this section, we first provide a detailed overview of the pretraining setup of Point-MGE. We then evaluate the pretrained model's performance in both representation learning and generation capabilities. Finally, to gain a deeper understanding of the roles and importance of each component in the model, we conduct ablation studies on Point-MGE. More results can be found in the Appendix.

\begin{table}[!t]
	\centering
	\footnotesize
	\setlength{\tabcolsep}{5pt}
	\begin{tabular}{lcccc}
		\toprule
		\multirow{2}{*}{Methods} &\multicolumn{2}{c}{5-way} &\multicolumn{2}{c}{10-way} \\
		\cmidrule(r){2-3} \cmidrule(r){4-5}
		&10-shot &20-shot &10-shot &20-shot\\
		\midrule
		\multicolumn{5}{c}{\textit{Supervised Learning Only}}\\
		\midrule
		PointNet &52.0±3.8 &57.8±4.9 &46.6±4.3 &35.2±4.8\\
		PointNet-OcCo &89.7±1.9 &92.4±1.6 &83.9±1.8 &89.7±1.5\\
		DGCNN &31.6±2.8 &40.8±4.6 &19.9±2.1 &16.9±1.5\\
		\midrule
		\multicolumn{5}{c}{\textit{with Single-Modal Pre-Training}}\\
		\midrule
		Point-BERT &94.6±3.1 &96.3±2.7 &91.0±5.4 &92.7±5.1\\
		Mask-Point &95.0±3.7 &97.2±1.7 &91.4±4.0 &93.4±3.5\\
		Point-MAE &96.3±2.5 &97.8±1.8 &92.6±4.1 &95.0±3.0\\
		Point-M2AE &96.8±1.8 &98.3±1.4 &92.3±4.5 &95.0±3.0 \\
		PointGPT &96.8±2.0 &98.6±1.1 &92.6±4.6 &95.2±3.4 \\
		GPM &97.2±2.6 &98.7±2.2 &\textbf{92.9±4.2} &95.0±3.0\\
		\textbf{Point-MGE} &\textbf{97.2±1.7} &\textbf{98.7±0.7} &92.8±4.0 &\textbf{95.4±3.2}\\
		\textit{Improvement} &\textit{+2.6} &\textit{+2.4} &\textit{+1.8} &\textit{+2.7}\\
		\midrule
		PointGPT-B &97.5±2.0 &98.8±1.0 &93.5±4.0 &\textbf{95.8±3.0} \\
		\textbf{Point-MGE-B} &\textbf{97.7±1.8} &\textbf{98.8±0.8} &\textbf{93.8±4.6} &95.8±3.4\\
		\midrule
		\multicolumn{5}{c}{\textit{with Cross-Modal Pre-Training}}\\
		\midrule
		ACT &96.8±2.3 &98.0±1.4 &93.3±4.0 &95.6±2.8\\
		Joint-MAE &96.7±2.2 &97.9±1.8 &92.6±3.7 &95.1±2.6\\
		I2P-MAE &97.0±1.8 &98.3±1.3 &92.6±5.0 &95.5±3.0\\
		Recon &97.3±1.9 &98.9±1.2 &93.3±3.9 &95.8±3.0\\
		\bottomrule
	\end{tabular}
	\caption{\textbf{Few-shot classification on ModelNet40~\protect\cite{Wu_2015_CVPR}.} We report the average accuracy (\%) and standard deviation (\%) of 10 independent experiments.}
	\label{Few-shot classification}
\end{table}

\subsection{Pre-training Setups}
Point-MGE is pre-trained on the ShapeNet~\cite{chang2015shapenet} dataset, which includes 55 different object categories and over 50,000 3D models. From the surface of each 3D shape, 2048 points are uniformly sampled and further divided into 64 patches, each containing 32 points. First, we trained the VQVAE to obtains the discrete tokens. Next, we determined the sliding masking ratio, setting $\beta=0.5$ and $u = 2$, and pre-trained the ViT-based extractor-generator architecture using multi-scale masked token reconstruction. The number of scales and the number of point patches at each scale were consistent with Point-M2AE~\cite{zhang2022point}. Additionally, to validate the high-capacity model learning ability of the Point-MGE framework, we scaled the model to ViT-B as described in \cite{NEURIPS2023_5ed5c3c8}, performed post-training, and evaluated its representation learning performance across various downstream tasks. We used the AdamW optimizer to perform pre-training for 300 epochs. Further implementation details and hyperparameter settings are provided in the Appendix.

\subsection{Representation Learning}
To demonstrate the representation learning performance of our method, we conducted experiments on shape classification, few-shot learning, and part segmentation. We evaluated the benchmark performance for supervised learning only, single-modal pre-training, and cross-modal pre-training separately.

\paragraph{Shape Classification.}
We conducted shape classification fine-tuning on two datasets. The ModelNet40~\cite{Wu_2015_CVPR} dataset includes 3D CAD models of 40 categories, and we followed the official data split provided in ~\cite{Wu_2015_CVPR} for our experiments. As shown in Table~\ref{Shape Classification}, Point-MGE achieved a second-place classification accuracy of 94.2\% on this dataset. The ScanObjectNN ~\cite{Uy_2019_ICCV}dataset, derived from real-world indoor scans, contains approximately 15,000 objects across 15 categories. We evaluated this dataset using three different data splits: OBJ-BG, OBJ-ONLY, and PB-T50-RS. The experimental results indicate that Point-MGE consistently outperformed other models across all splits. Notably, in the OBJ-BG setting, Point-MGE achieved an accuracy of 92.9\%, outperforming the second-best pre-trained model~\cite{NEURIPS2023_5ed5c3c8} by 1.3 percentage points. Additionally, we extended Point-MGE and three benchmark methods to the ViT-B model for performance evaluation. The results demonstrate that under high-capacity model conditions, our method still maintains a competitive advantage.

\begin{table}[!t]
	\centering
	\footnotesize
	\setlength{\tabcolsep}{5pt}
	\begin{tabular}{lcc}
		\toprule
		Methods & $\text{mIoU}_C$~$\uparrow$ & $\text{mIoU}_I$~$\uparrow$\\
		\midrule
		\multicolumn{3}{c}{Supervised Learning Only}\\
		\midrule
		PointNet~\cite{Qi_2017_CVPR}  &80.4 &83.7\\
		PointNet++~\cite{qi2017pointnet++}  &81.9 &85.1\\ 
		DGCNN~\cite{wang2019dynamic}  &82.3 &85.2\\
		\midrule
		\multicolumn{3}{c}{with Single-Modal Pre-Training}\\
		\midrule
		Transformer + OcCo~\cite{Yu_2022_CVPR}  &83.4 &85.1\\
		Point-BERT~\cite{Yu_2022_CVPR}  &84.1 &85.6\\
		Mask-Point~\cite{liu2022masked}  &84.4 &86.0\\
		Point-MAE~\cite{pang2022masked}    &- &86.1\\
		PointGPT~\cite{NEURIPS2023_5ed5c3c8}  &84.1 &86.2\\
		GPM~\cite{li2024general} &84.2 &85.8 \\
		\textbf{Point-MGE} &\textbf{85.0} &\textbf{86.5}\\
		\textit{Improvement} &\textit{+0.9} &\textit{+0.9}\\
		\midrule
		PointGPT-B~\cite{NEURIPS2023_5ed5c3c8}  &84.5 &86.5\\
		\textbf{Point-MGE-B} &\textbf{85.0} &\textbf{86.6}\\
		\midrule
		\multicolumn{3}{c}{with Cross-Modal Pre-Training}\\
		\midrule
		ACT~\cite{dong2023autoencoders}  &84.7 &86.1\\
		Recon~\cite{qi2023contrast} &84.8 &86.4\\
		\bottomrule
	\end{tabular}
	\caption{\textbf{Part segmentation on ShapeNetPart~\protect\cite{yi2016scalable}.} We report the mean IoU (\%) of all part categories ('$\text{mIoU}_C$') and the mean IoU of all instances ('$\text{mIoU}_I$') in the dataset.}
	\label{Part segmentation}
\end{table}

\paragraph{Few-shot Classification.}
We evaluated the few-shot learning performance of Point-MGE by conducting N-way, K-shot experiments on the ModelNet40\cite{Wu_2015_CVPR} dataset. Specifically, we randomly selected N classes from ModelNet40 and sampled K objects from each class. Each experiment was repeated 10 times, and the average and standard deviation of the results were calculated. We evaluated each of the four combinations of settings with N = \{5, 10\} and K = \{10, 20\}. As shown in Table \ref{Few-shot classification}, under both ViT and ViT-B settings, we achieved SOTA performance in at least three combinations, demonstrating that Point-MGE's pre-trained 3D representations are highly effective for few-shot learning tasks.

\paragraph{Part Segmentation.}
Table~\ref{Part segmentation} presents the results of average IoU for all categories ($\text{mIoU}_{C}$) and all instances ($\text{mIoU}_{I}$) in the part segmentation experiments on the ShapeNetPart~\cite{yi2016scalable} dataset. Compared to the fully supervised method DGCNN\cite{wang2019dynamic}, Point-MGE achieved significant improvements of 2.7\% and 1.3\% in these two key evaluation metrics, respectively. Our method also outperforms other pretrained models, achieving a $\text{mIoU}_C$ of 85.0\%, which ranks first. Additionally, the experimental results show that increasing the number of parameters yields only marginal improvements for this task.

\subsection{3D Shape Generation}

\begin{table}[!t]
	\centering
	\footnotesize
	\setlength{\tabcolsep}{4pt}
	\begin{tabular}{lcccc}
		\toprule
		\multirow{2}{*}{Methods} & \multirow{2}{*}{S-FPD~$\downarrow$} &S-KPD~$\downarrow$  & \multirow{2}{*}{R-FID~$\downarrow$} &R-KID~$\downarrow$\\
		& & ($\times10^3$) & &($\times10^3$)\\
		\midrule
		Grid-$8^3$  &4.03 &6.15 &32.78 &14.12\\
		PVD &2.33 &2.65 &270.64 &281.5 \\
		3DILG &1.89 &2.17 &24.83 &\textbf{10.51}\\
		3DShape2VecSet &0.97 &\textbf{1.11} &24.24 &11.76\\ 
		\textbf{Point-MGE} &\textbf{0.85} &1.34 &\textbf{21.63} &11.48\\
		\bottomrule
	\end{tabular}
	\caption{\textbf{Unconditional generation.} We report four metrics for evaluating the unconditional shape generation task.}
	\label{Unconditional Generation}
\end{table}

\begin{figure}[!t]
	\centering
	\includegraphics[width=1.0\linewidth,keepaspectratio]{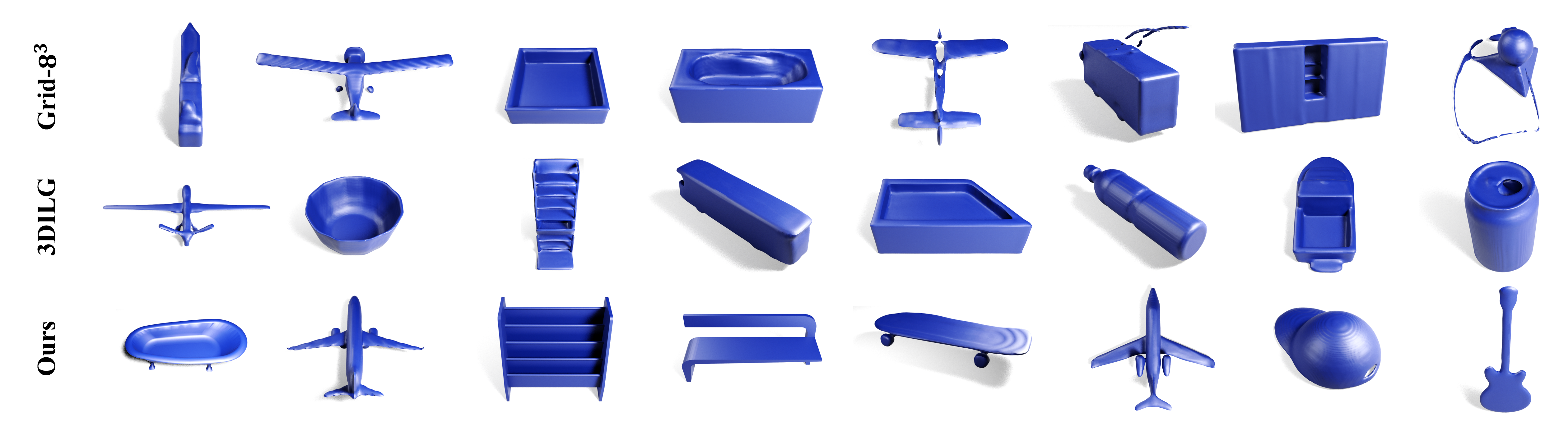}
	\caption{\textbf{Visualization of unconditional generation.} All models were trained on the complete ShapeNet dataset, and we sampled nine generated results for rendering.}
	\label{uncond vis}
\end{figure}

\paragraph{Unconditional Generation.}
After completing the pre-training stage, Point-MGE is capable of performing unconditional generation tasks. fréchet inception distance (FID) and kernel inception distance (KID) are commonly used to evaluate the generation quality of generative models. Following~\cite{zhang20233dshape2vecset}, we use Rendering-FID (R-FID) and Rendering-KID (R-KID) metrics to evaluate the FID and KID of rendered images of 3D shapes, as well as Surface-FPD (S-FPD) and Surface-KPD (S-KPD) metrics to evaluate the FID and KID of sampled point clouds from 3D shapes. Table~\ref{Unconditional Generation} compares the performance of various methods in unconditional shape generation on ShapeNet. 3DILG~\cite{zhang20223dilg} also adopts a two-stage generation strategy and completes the generation training through an autoregressive process. PVD~\cite{zhou20213d} and 3DShape2VecSet~\cite{zhang20233dshape2vecset} are generative methods based on diffusion models. Point-MGE outperforms other methods on two of the evaluation metrics, indicating that our approach can generate 3D shapes that closely match the real distribution. Additionally, in Figure~~\ref{uncond vis}, we visualize the rendered results of the generated shapes.

\begin{table}[!t]
	\centering
	\footnotesize
	\setlength{\tabcolsep}{3.5pt}
	\begin{tabular}{l|lcccc}
		\toprule
		&Methods &Airplane  &Faucet  &Table  &Chair\\
		\midrule
		\multirow{4}{*}{S-FID~$\downarrow$}
		&3DILG &0.71 &2.49  &2.10 &0.96\\
		&NeuralWavelet &\textbf{0.38} &1.84  &1.12 &1.14\\
		&3DShape2VecSet &0.62 &\textbf{1.47} &1.19 &0.76\\
		&\textbf{Point-MGE} &0.67 &1.65  &\textbf{1.03} &\textbf{0.71} \\
		\midrule
		\multirow{4}{*}{\makecell{S-KID~$\downarrow$\\($\times 10^3$)}}
		&3DILG &0.81 &3.76  &3.84 &1.21\\
		&NeuralWavelet &\textbf{0.53} &2.41  &1.55 &1.50\\
		&3DShape2VecSet &0.83 &1.34  &1.87 &0.70\\
		&\textbf{Point-MGE} &0.79 &\textbf{1.31}  &\textbf{1.14} &\textbf{0.58}\\
		\bottomrule
	\end{tabular}
	\caption{\textbf{Category-conditional generation.} We reported the S-FID and S-KID metrics for four categories.}
	\label{cls-Generation}
\end{table}

\begin{figure}[!t]
	\centering
	\includegraphics[width=1.0\linewidth,keepaspectratio]{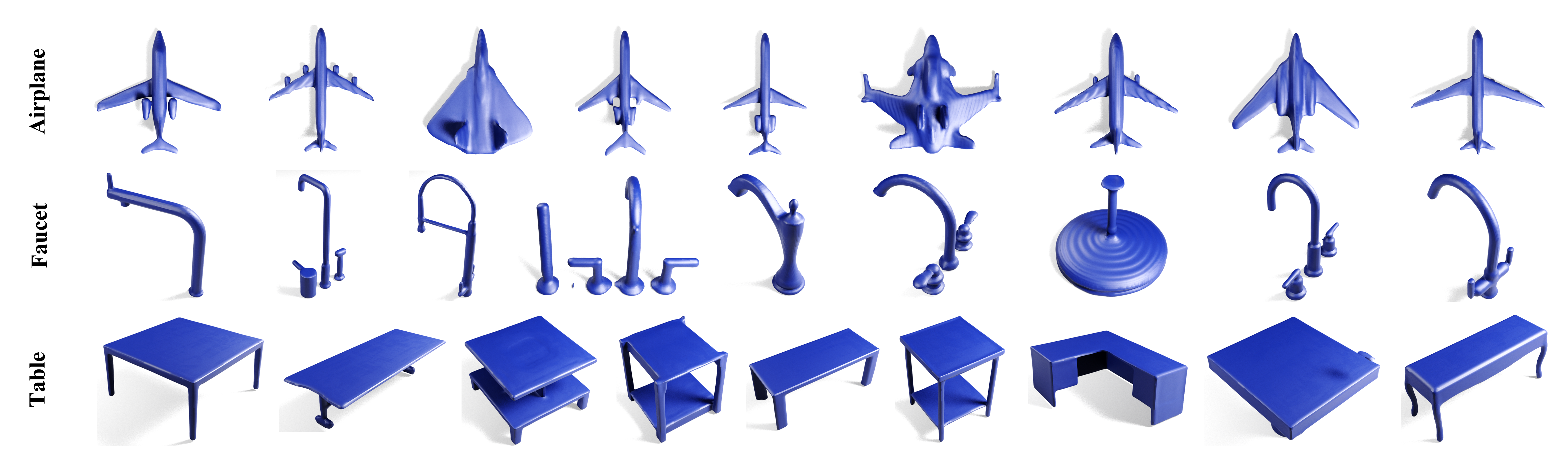}
	\caption{\textbf{Visualization of category-conditional generation.} We selected three category labels (airplane, faucet, and table) to display the rendered results of the generated shapes.}
	\label{cls-cond vis}
\end{figure}

\begin{figure}[!t]
	\centering
	\includegraphics[width=1.0\linewidth,keepaspectratio]{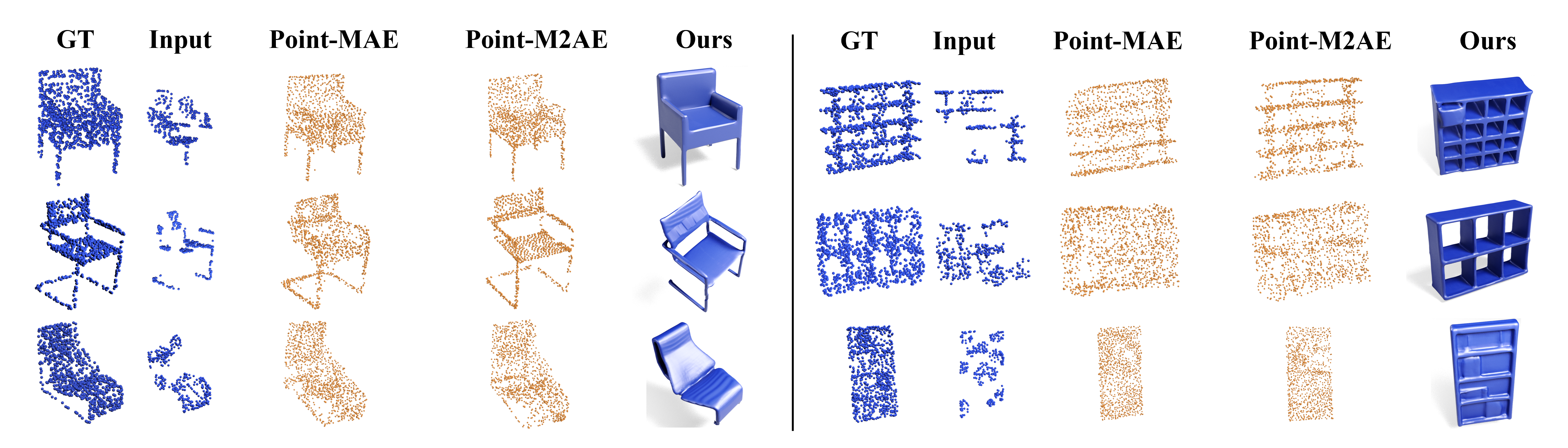}
	\caption{\textbf{Visualization of masked point clouds conditional generation.} We used a block masking strategy to mask 70\% of the complete point cloud as conditional input. The generated results demonstrated the ability of different models to reconstruct the complete shapes.}
	\label{point-cond vis}
\end{figure}

\begin{table}[!t]
	\centering
	\footnotesize
	\setlength{\tabcolsep}{12pt}
	\begin{tabular}{cccc}
		\toprule
		Extractor &Generator  &Acc (\%)        &R-FID~$\downarrow$ \\
		\midrule
		6         &6          &91.2            &29.07 \\
		8         &6          &91.9            &27.85 \\
		\rowcolor{blue!8}
		12        &8          &\textbf{92.4}   &21.63 \\
		12        &12         &92.1            &\textbf{20.94} \\
		\bottomrule
	\end{tabular}
	\caption{\textbf{Effects of varying transformer depths.} The accuracy (\%) on the ScanObjectNN~\protect\cite{Uy_2019_ICCV} (OBJ\_BG) and the R-FID for unconditional generation are reported.}
	\label{Ablation Transformer}
\end{table}

\paragraph{Conditional Generation.}
We evaluated the conditional generation capabilities of Point-MGE under two different settings. First, we retrained the model using the 54 categories from the ShapeNet~\cite{chang2015shapenet} dataset as conditional embeddings. As shown in Table~\ref{cls-Generation}, we evaluated the Surface-FID (S-FID) and Surface-KID (S-KID) for our method and the benchmark methods across four categories. The results indicate that our method achieved performance comparable to or even surpassing those of purely generative approaches. Figure~\ref{cls-cond vis} displays the rendered results of generated shapes for three categories (airplane, faucet, and table), with nine instances shown for each category. Next, we further evaluated Point-MGE against other SSL models in the masked generation task. As shown in Figure~\ref{point-cond vis}, these models typically train by reconstructing local point clouds, which primarily allows them to complete point cloud reconstruction tasks and depends on the provision of the central point positions. This approach, to some extent, leaks complete shape information to the decoder. In contrast, our method can reconstruct 3D shape meshes using the minimal amount of point cloud information and does not rely on masked center point information during the generation process.

\begin{table}[!t]
	\centering
	\footnotesize
	\setlength{\tabcolsep}{13pt}
	\begin{tabular}{ccc}
		\toprule
		Type                    &Acc (\%)       &R-FID~$\downarrow$ \\
		\midrule
		\rowcolor{blue!8}
		Quantized features      &\textbf{92.4}   &21.63 \\
		Index-embedded tokens   &87.2            &\textbf{20.34}\\
		\bottomrule
	\end{tabular}
	\caption{\textbf{Effects of input type.} The accuracy (\%) and the R-FID are reported.}
	\label{Ablation input}
\end{table}

\begin{table}[!t]
	\centering
	\footnotesize
	\setlength{\tabcolsep}{3pt}
	\begin{tabular}{l|ccccc}
		\toprule
		& \multicolumn{3}{c}{Sliding Masking}& \multicolumn{2}{c}{Various Masking}\\
		\cmidrule(r){2-4} \cmidrule(r){5-6}
		&$\beta=0.5$   &\cellcolor{blue!8}$\beta=0.5$  &$\beta=0.7$  &$\mu=1.0$   &$\mu=0.7$\\
		&$u=1$  &\cellcolor{blue!8}$u=2$ &$u=2$  &$\sigma=0$ &$\sigma=0.25$ \\
		\midrule
		Acc (\%)   &91.9       &\cellcolor{blue!8}\textbf{92.4}       &91.7      &86.2       &91.6 \\
		R-FID~$\downarrow$  &24.25    &\cellcolor{blue!8}\textbf{21.63} &22.71      &26.72    &22.43\\
		\bottomrule
	\end{tabular}
	\caption{\textbf{Effects of masking design.}The accuracy (\%) and the R-FID are reported.}
	\label{Ablation masking}
\end{table}

\begin{table}[!t]
	\centering
	\footnotesize
	\setlength{\tabcolsep}{10pt}
	\begin{tabular}{lcccc}
		\toprule
		&\cellcolor{blue!8}NeRFs    &EMD &CD L1 &CD L2 \\
		\midrule
		Acc (\%)      &\cellcolor{blue!8}\textbf{92.4}   &87.9 &88.7 &89.1 \\
		R-FID~$\downarrow$ &\cellcolor{blue!8}21.63 &N/A  &N/A  &N/A \\
		\bottomrule
	\end{tabular}
	\caption{\textbf{Effects of input type.} The accuracy (\%) and the R-FID are reported.}
	\label{Ablation Reconstraction}
\end{table}

\subsection{Ablation Studies}
In this section, we discuss several factors that influence the representation learning and generative capabilities of the pre-trained model.

\paragraph{Transformer Depth.}
As shown in Table~\ref{Ablation Transformer}, we used the standard single scale ViT architecture to build the extractor-generator structure and explored four different transformer depth combinations. The results indicate that increasing the model depth significantly enhances the performance of the pre-trained model. Additionally, consistent with trends observed in previous masked modeling studies~\cite{he2022masked,pang2022masked,zhang2022point}, an asymmetric extractor-generator structure is more beneficial for representation learning. This reliance on the extractor to obtain high-quality representations from a small number of visible tokens is due to the use of a relatively simple generator to complete the reconstruction task.

\paragraph{Input Type.}
In Table~\ref{Ablation input}, we investigate the impact of two different input types on model performance. Specifically, after vector quantizing the initial features of the point cloud patches, we consider using either the quantized features or their indices in the codebook as inputs. Experimental results show that using index-embedded tokens benefits the generation task, but its generalization ability across datasets is limited, which is detrimental to fine-tuning on downstream tasks, leading to poorer representation learning capabilities.

\paragraph{Masking Design.}
In determining the optimal masking strategy, as shown in Table~\ref{Ablation masking}, we compared our sliding masking strategy with the variable masking strategy used in MAGE~\cite{Li_2023_CVPR} and analyzed the impact of different hyperparameters on performance. For the variable masking, $\mu$ and $\sigma$ represent the mean and standard deviation of the truncated Gaussian distribution, respectively. The results indicate that the variable masking ratio, which performs well on 2D images, does not provide an advantage in the point cloud modality, where data is relatively scarce. In contrast, our smooth masking strategy better facilitates the transition between representation learning and generative training.

\paragraph{Reconstruction Target.}
In exploring the training of VQVAE, we compared different reconstruction targets. As shown in Table 9, this study employs NeRFs as the reconstruction target. Additionally, we experimented with directly using point clouds as the reconstruction target and tested methods using Earth Mover's Distance (EMD) and Chamfer Distance (CD) as loss functions. The experimental results demonstrate that our approach excels in overcoming point cloud sampling bias, leading to superior representation learning capabilities. However, directly using point clouds as the reconstruction target, similar to previous methods, is limited to generating point clouds and cannot produce high-resolution 3D shapes. In the table, this generative capability is marked as Not Applicable (N/A).

\section{Conclusion}
We present Point-MGE, a novel unified framework for point cloud shape generation and representation learning. Our approach serializes point patches to obtain quantized tokens and applies a sliding masking ratio, allowing it to adapt to both representation learning and generation tasks. Our method excels in addressing the sampling bias issue in point clouds, setting new SOTA results in shape classification, few-shot learning, and part segmentation tasks. Additionally, in generation tasks, our approach achieves near SOTA performance in both unconditional and conditional shape generation, producing realistic high-resolution 3D shapes. We hope our work will draw attention from researchers to the synergistic effects of combining these two model types in 3D vision.

\newpage
\clearpage
\section*{Acknowledgments}
This work was supported by the Guangdong Major Project of  Basic and Applied Basic Research (2023B0303000016).

\bibliographystyle{named}
\bibliography{PointMGE}

\end{document}